\documentclass{article}

\usepackage[final]{dlde_neurips_2022}

\usepackage[utf8]{inputenc}             
\usepackage[T1]{fontenc}                
\usepackage[backref=section]{hyperref}  
\usepackage{bookmark}
\usepackage{url}                        
\usepackage{booktabs}                   
\usepackage{amsfonts}                   
\usepackage{nicefrac}                   
\usepackage{microtype}                  
\usepackage[dvipsnames]{xcolor}         

\usepackage[frozencache]{minted} 
\hypersetup{
  breaklinks,
  colorlinks,
  linkcolor=BrickRed,
  citecolor=MidnightBlue,
  urlcolor=MidnightBlue,
}

\usepackage{graphicx}
\usepackage{tikz}
\usepackage{caption}
\usepackage{subcaption}
\usepackage{wrapfig}
\usepackage{import} 

\usepackage{amsmath}
\usepackage{amssymb}
\usepackage{bm}
\usepackage{mathtools}

\usepackage{siunitx}
\usepackage{etoolbox}
\renewrobustcmd{\bfseries}{\fontseries{b}\selectfont}
\renewrobustcmd{\boldmath}{}

\usepackage[shortcuts=abbr]{glossaries-extra}
\glsdisablehyper{}

\newcommand\extrafootertext[1]{%
    \bgroup%
    \renewcommand\thefootnote{\fnsymbol{footnote}}%
    \renewcommand\thempfootnote{\fnsymbol{mpfootnote}}%
    \footnotetext[0]{#1}%
    \egroup%
}

\usepackage{enumitem}
\setlist[itemize]{noitemsep,left=7pt,nosep}

\newcommand\package[1]{\texttt{#1}}
\def\torchode{\package{torchode}}
\def\torchdiffeq{\package{torchdiffeq}}
\def\diffrax{\package{diffrax}}
\def\torchdyn{\package{TorchDyn}}
\def\diffeq{\package{DifferentialEquations.jl}}
\def\diffeqflux{\package{DiffEqFlux.jl}}

\def\python{Python}
\def\julia{Julia}
\def\pytorch{PyTorch}
\def\jax{JAX}

\usepackage{pifont}
\newcommand{\cmark}{\ding{51}}%
\newcommand{\xmark}{\ding{55}}%
\def\yes{\cmark}
\def\no{\xmark}

\newcommand{\defabbrcmd}[1]{
\expandafter\def\csname#1\endcsname{{\ab{#1}}}
}
\newcommand{\defabbrcmds}[1]{
\expandafter\def\csname#1\endcsname{{\ab{#1}}}
\expandafter\def\csname#1s\endcsname{{\abp{#1}}}
}

\newabbreviation{ODE}{ODE}{ordinary differential equation}
\defabbrcmds{ODE}

\newabbreviation{ml}{ML}{machine learning}
\defabbrcmds{ml}

\newabbreviation{cnf}{CNF}{continous normalizing flow}
\defabbrcmds{cnf}

\newabbreviation{fen}{FEN}{finite element networks}
\defabbrcmd{fen}

\newabbreviation{vdp}{VdP}{Van der Pol}
\defabbrcmd{vdp}

\newabbreviation{jit}{JIT}{just-in-time}
\defabbrcmd{jit}

\newcommand\py[1]{\mintinline{python}{#1}}
\newcommand\version[1]{\texttt{#1}}

\usepackage[capitalise]{cleveref}

\title{torchode: A Parallel ODE Solver for PyTorch}

\author{%
  Marten Lienen \& Stephan G\"unnemann \\
  Department of Informatics \& Munich Data Science Institute \\
  Technical University of Munich, Germany \\
  \{\texttt{m.lienen,s.guennemann}\}\texttt{@tum.de}
}

\begin{document}

\maketitle

\begin{abstract}\makeatletter\phantomsection\def\@currentlabel{abstract}\makeatother
  We introduce an ODE solver for the \pytorch{} ecosystem that can solve multiple ODEs in parallel independently from each other while achieving significant performance gains. Our implementation tracks each ODE's progress separately and is carefully optimized for GPUs and compatibility with \pytorch{}'s JIT compiler. Its design lets researchers easily augment any aspect of the solver and collect and analyze internal solver statistics. In our experiments, our implementation is up to 4.3 times faster per step than other ODE solvers and it is robust against within-batch interactions that lead other solvers to take up to 4 times as many steps.
\end{abstract}

\section{Introduction}\label{sec:introduction}

\extrafootertext{Code is available at \href{https://github.com/martenlienen/torchode}{github.com/martenlienen/torchode}.}

\Abp{ODE} are the natural framework to represent continuously evolving systems. They have been applied to the continuous transformation of probability distributions \citep{chen2018neural,grathwohl2019ffjord}, modeling irregularly-sampled time series \citep{debrouwer2019gruodebayes,rubanova2019latent}, and graph data \citep{poli2019graph} and connected to numerical methods for PDEs \citep{lienen2022learning}. Various extensions \citep{dupont2019augmented,xia2021heavy,norcliffe2021neural} and regularization techniques \citep{pal2021opening,ghosh2020steer,finlay2020how} have been proposed and \citep{gholami2019anode,massaroli2020dissecting,ott2021resnet} have analyzed the choice of hyperparameters and model structure. Despite the large interest in these methods, the performance of \pytorch{} \citep{paszke2019pytorch} \ODE{} solvers has not been a focus point and benchmarks indicate that solvers for \pytorch{} lag behind those in other ecosystems.\footnote{\href{https://benchmarks.sciml.ai}{benchmarks.sciml.ai}, \href{https://github.com/patrick-kidger/diffrax/tree/main/benchmarks}{github.com/patrick-kidger/diffrax/tree/main/benchmarks}}

\torchode{} aims to demonstrate that faster model training and inference with \ODEs{} is possible with \pytorch{}. Furthermore, parallel, independent solving of batched \ODEs{} eliminates unintended interactions between batched instances that can dramatically increase the number of solver steps and introduce noise into model outputs and gradients.

\section{Related Work}\label{sec:related-work}

The most well-known \ODE{} solver for \pytorch{} is \torchdiffeq{} that popularized training with the adjoint equation \citep{chen2018neural}. Their implementation comes with many low- to medium-order explicit solvers and has been the basis for a differentiable solver for controlled differential equations \citep{kidger2020neural}. Another option in the \pytorch{} ecosystem is \torchdyn{}, a collection of tools for implicit models that includes an \ODE{} solver but also utilities to plot and inspect the learned dynamics \citep{poli2021torchdyn}. \torchode{} goes beyond their \ODE{} solving capabilities by solving multiple independent problems in parallel with separate initial conditions, integration ranges and solver states such as accept/reject decisions and step sizes, and a particular concern for performance such as compatibility with \pytorch{}'s \jit{} compiler.

Recently, \citeauthor{kidger2022neural} has released with \diffrax{} (\citeyear{kidger2022neural}) a collection of solvers for \ODEs{}, but also controlled, stochastic, and rough differential equations for the up-and-coming deep learning framework \jax{} \citep{bradbury2018jax}. They exploit the features of \jax{} to offer many of the same benefits that \torchode{} makes available to the \pytorch{} community and \diffrax{}'s internal design was an important inspiration for the structure of our own implementation.

Outside of \python{}, the \julia{} community has an impressive suite of solvers for all kinds of differential equations with \diffeq{} \citep{rackauckas2017differentialequations}. After a first evaluation of different types of sensitivity analysis in \citeyear{ma2018comparison} (\citeauthor{ma2018comparison}), they released \diffeqflux{} which combines their \ODE{} solvers with a popular deep learning framework \citep{rackauckas2019diffeqflux}.

\section{Design \& Features of \torchode}\label{sec:design}

We designed \torchode{} to be correct, performant, extensible and introspectable. The former two aspects are, of course, always desirable, while the latter two are especially important to researchers who may want to extend the solver with, for example, learned stepping methods or record solution statistics that the authors did not anticipate.

\setlength{\intextsep}{0pt}
\begin{wraptable}[9]{r}{9.5cm}
  \centering
  \caption{Feature comparison with existing \pytorch{} \ODE{} solvers.}\label{table:features}
  \vspace{-3pt}
  \begin{tabular}{rccccc}
    \toprule
    {} & \torchode{} & \torchdiffeq{} & \torchdyn{} \\
    \midrule
    Parallel solving & \yes{} & \no{} & \no{} \\
    JIT compilation & \yes{} & \no{} & \no{} \\
    Extensible & \yes{} & \no{} & \yes{} \\
    Solver statistics & \yes{} & \no{} & \no{} \\
    Step size controller & PID & I & I \\
    \bottomrule
  \end{tabular}
\end{wraptable}
The major architectural difference between \torchode{} and existing \ODE{} solvers for \pytorch{} is that we treat the batch dimension in batch training explicitly (\cref{table:features}). This means that the solver holds a separate state for each instance in a batch, such as initial condition, integration bounds and step size, and is able to accept or reject their steps independently. Batching instances together that need to be solved over different intervals, even of different lengths, requires no special handling in \torchode{} and even parameters such as tolerances could be specified separately for each problem. Most importantly, our parallel integration avoids unintended interactions between problems in a batch that we explore in \cref{sec:batching}.

Two other aspects of \torchode{}'s design that are of particular importance in research are extensibility and introspectability. Every component can be re-configured or easily replaced with a custom implementation. By default, \torchode{} collects solver statistics such as the number of total and accepted steps. This mechanism is extensible as well and lets a custom step size controller, for example, return internal state to the user for further analysis without relying on global state.

The speed of model training and evaluation constrains computational resources as well as researcher productivity. Therefore, performance is a critical concern for \ODE{} solvers and \torchode{} takes various implementation measures to optimize throughput as detailed below and evaluated in \cref{sec:benchmarks}. Another way to save time is the choice of time step. It needs to be small enough to control error accumulation but as large as possible to progress quickly. \torchode{} includes a PID controller that is based on analyzing the step size problem in terms of control theory \citep{soderlind2002automatic,soderlind2003digital}. These controllers generalize the integral (I) controllers used in \torchdiffeq{} and \torchdyn{} and are included in \diffeq{} and \diffrax{}. In our evaluation in \cref{sec:pid} these controllers can save up to 5\% of steps if the step size changes quickly.

\paragraph{What makes \torchode{} fast?} \ODE{} solving is inherently sequential except for efforts on parallel-in-time solving \citep{gander201550}. Taking the evaluation time of the dynamics as fixed, performance of an \ODE{}-based model can therefore only be improved through a more efficient implementation of the solver's looping code, so as to minimize the time between consecutive dynamics evaluations. In addition to the common FSAL and SSAL optimizations for Runge-Kutta methods to reuse intermediate results, \torchode{} avoids expensive operations such as conditionals evaluated on the host that require a CPU-GPU synchronization as much as possible and seeks to minimize the number of \pytorch{} kernels launched. We rely extensively on operations that combine multiple instructions in one kernel such as \py{einsum} and \py{addcmul}, in-place operations, pre-allocated buffers, and fast polynomial evaluation via Horner's rule that saves half of the multiplications over the naive evaluation method. Finally, \jit{} compilation minimizes \python{}'s CPU overhead and allows us to reach even higher GPU utilization.

\paragraph{What slows \torchode{} down?} The extra cost incurred by tracking a separate solver state for every problem is negligible on a highly parallel computing device such as a GPU.\@ However, because each \ODE{} progresses at a different pace, they might pass a different number of evaluation points at each step. Keeping track of this requires indexing with a Boolean tensor, a relatively expensive operation.

\section{Experiments}\label{sec:experiments}

\subsection{Batching ODEs: What could possibly go wrong?}\label{sec:batching}

\begin{figure}[tbp]
  \centering
  \includegraphics{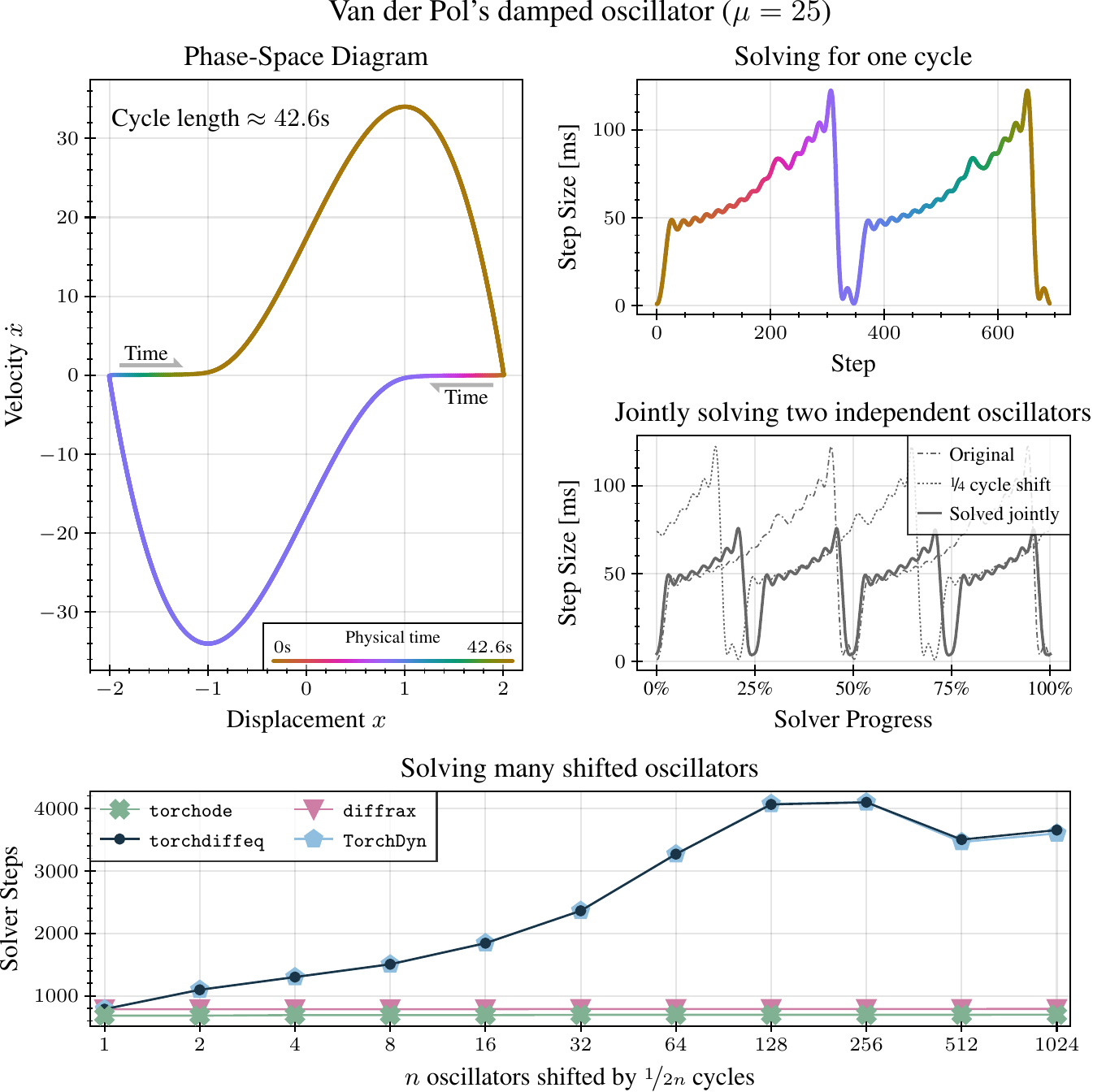}
  \caption{Van der Pol's oscillator is a cyclic system with nonlinear damping that exhibits a strong variation in step size under explicit methods such as 5th order Dormand-Prince. If multiple oscillators are treated jointly, the stiffest oscillator dominates the common step size, increasing the number of solver steps unnecessarily. \torchode{} solves the instances independently, keeping the steps constant and the efficiency high. Step sizes have been smoothed by removing high-frequency variations.}\label{fig:van-der-pol}
\end{figure}

As is established practice in deep learning, mini-batching of instances is also common in the training of and inference with neural \ODEs{}. A mini-batch is constructed by concatenating a set of $n$ initial value problems of size $p$ and then solving it as a single problem of size $np$. Since the learned dynamics still apply to each instance independently, this should have no adverse effects. However, jointly solving the individual problems means that they share step size and error estimate, and solver steps will be either accepted or rejected for all instances at once. In effect, the solver tolerances for a certain initial value problem vary depending on the behavior of the other problems in the batch.

To investigate this problem, we will consider a damped oscillator as in \ab{vdp}['s] equation
\begin{equation}
  \ddot{x} = \mu(1 - x^{2})\dot{x} - x. \label{eq:vdp}
\end{equation}
If the damping $\mu$ is significantly greater than $0$, \cref{eq:vdp} has time-varying stiffness which means that an explicit solver (as is commonly used with neural \ODEs{}) will exhibit a significant variation in step size over the course of a cycle of the oscillator. If we combine multiple instances of the oscillator with varying initial conditions in a batch, the common step size of the batch at any point in time will be roughly the minimum over the step sizes of the individual instances. Therefore, \torchdiffeq{} and \torchdyn{} need up to four times as many steps to solve a batch of these problems as the parallel solvers of \torchode{} and \diffrax{}. See \cref{fig:van-der-pol} for a visual explanation of the phenomenon.

While the scenario of stacked \vdp{} problems mainly reduces the efficiency of the solver, we believe that one could also construct an ``adversarial'' example that maximizes the error of a specific instance in a batch by controlling its effective tolerances.

\subsection{Benchmarks}\label{sec:benchmarks}

We evaluate \torchode{} against \torchdiffeq{} and \torchdyn{} in three settings: solving the \vdp{} equation and two learning scenarios to measure the impact that a carefully tuned, parallel implementation can have on training and inference of \ml{} models. First, we consider \fen{}, a graph neural network that learns the dynamics of physical systems \citep{lienen2022learning}, which we train via backpropagation through the solver (discretize-then-optimize). Second, we consider a \cnf{} based on the FFJORD method \citep{grathwohl2019ffjord}, which is trained via the adjoint equation (optimize-then-discretize) \citep{chen2018neural}.

The results in \cref{table:benchmark} show that \torchode{}'s solver loop is significantly faster than \torchdiffeq{}'s. Additionally, \jit{} compilation roughly doubles \torchode{}'s throughput. See \cref{sec:benchmarks--details} for the full results, a detailed discussion and a complete description of the setups. The independent solving of batch instances explored in \cref{sec:batching} seems to have only a small effect on the number of solver steps and achieved loss values (see \cref{sec:benchmarks--details}) for \fen{} and \cnf{}, most likely because, overall, the learned models exhibit only small variations in stiffness.

\begin{table}[t]
\centering
  \caption{Loop time (LT) in milliseconds (one solver step excluding model evaluation time) and corresponding speed up (SU) over \torchdiffeq{} on an \ODE{} (\vdp{}), a discretize-then-optimize (\fen{}) and an optimize-then-discretize setup with forward (\cnf{}-Fw.) and backward (\cnf{}-Bw.) solve.}\label{table:benchmark}
  \begin{tabular}{rccccccccc}
    \toprule
    {} & \multicolumn{2}{c}{\vdp{}} & \multicolumn{2}{c}{\fen{}} & \multicolumn{2}{c}{\cnf{}-Fw.} & \multicolumn{2}{c}{\cnf{}-Bw.} \\
    \cmidrule(lr){2-3}\cmidrule(lr){4-5}\cmidrule(lr){6-7}\cmidrule(lr){8-9}
    {} & LT & SU & LT & SU & LT & SU & LT & SU \\
    \midrule
    \torchdiffeq{} & 3.58 & \texttimes{}1.0 & 3.9 & \texttimes{}1.0 & 3.4 & \texttimes{}1.0 & 7.4 & \texttimes{}1.0 \\
    \torchdyn{} & 3.54 & \texttimes{}1.0 & 1.49 & \texttimes{}2.6 & 1.63 & \texttimes{}2.1 & 7.6 & \texttimes{}1.0 \\
    \torchode{} & 3.21 & \texttimes{}1.1 & 1.71 & \texttimes{}2.3 & \textbf{1.5} & \texttimes{}\textbf{2.3} & \textbf{2.38} & \texttimes{}\textbf{3.1} \\
    \torchode{}-\jit{} & \textbf{1.63} & \texttimes{}\textbf{2.2} & \textbf{0.91} & \texttimes{}\textbf{4.3} & - & - & - & - \\
    \bottomrule
  \end{tabular}
\end{table}

\section{Conclusion}\label{sec:conclusion}

We have shown that significant efficiency gains in the solver loop of continuous-time models such as neural \ODEs{} and \cnfs{} are possible. \torchode{} solves \ODE{} problems up to 4\texttimes{} faster than existing \pytorch{} solvers, while at the same time sidestepping any possible performance pitfalls and unintended interactions that can result from naive batching. Because \torchode{} is fully \jit{}-compatible, models can be \jit{} compiled regardless of where in their architecture they rely on \ODEs{} and automatically benefit from any future improvements to \pytorch{}'s \jit{} compiler. Finally, \torchode{} simplifies high-performance deployment of \ODE{} models trained with \pytorch{} by allowing them to be exported via ONNX (that relies on \jit{}) and run with an optimized inference engine such as \href{https://onnxruntime.ai}{onnxruntime}.

\bibliography{main}
\bibliographystyle{iclr2022_conference}

\clearpage
\appendix

\bookmarksetupnext{level=part}
\pdfbookmark{Appendix}{appendix}

\section{Detailed Benchmark Descriptions and Results}\label{sec:benchmarks--details}

\begin{table}[t]
\centering
  \caption{\vdp{}-Benchmarks. All times are measured in milliseconds.}\label{table:benchmark-vdp}
  \begin{tabular}{rccccc}
    \toprule
    {} & \torchode{} & \torchode{}-JIT & \torchdiffeq{} & \torchdyn{} & \diffrax{} \\
    \midrule
    loop time & \num{3.21+-0.11} & \num{1.629+-0.010} & \num{3.58+-0.04} & \num{3.54+-0.07} & \num{0.9014+-0.0011} \\
    \bottomrule
  \end{tabular}
\end{table}
\begin{table}[t]
\centering
  \caption{\fen{}-Benchmarks. All times are measured during the forward pass in milliseconds.}\label{table:benchmark-fen}
  \begin{tabular}{rcccc}
    \toprule
    {} & \torchode{} & \torchode{}-JIT & \torchdiffeq{} & \torchdyn{} \\
    \midrule
    loop time & \num{1.71+-0.05} & \num{0.91+-0.03} & \num{3.9+-0.3} & \num{1.49+-0.06} \\
    \midrule
    total time / step & \num{11.9+-0.3} & \num{10.92+-0.14} & \num{14.1+-0.4} & \num{11.2+-0.4} \\
    model time / step & \num{10.1+-0.3} & \num{9.92+-0.14} & \num{10.9+-0.3} & \num{9.6+-0.3} \\
    steps & \num{13.2+-0.2} & \num{13.3+-0.2} & \num{13.6+-0.2} & \num{13.3+-0.3} \\
    MAE & \num{0.845+-0.003} & \num{0.847+-0.005} & \num{0.846+-0.004} & \num{0.846+-0.004} \\
    \bottomrule
  \end{tabular}
\end{table}
\begin{table}[t]
\centering
  \caption{\cnf{}-Benchmarks. All times are measured in milliseconds.}\label{table:benchmark-cnf}
  \begin{tabular}{rcccc}
    \toprule
    {} & \torchode{} & \torchode{}-joint & \torchdiffeq{} & \torchdyn{} \\
    \midrule
    fw.\ loop time & \num{1.33+-0.16} & \num{1.5+-0.1} & \num{3.4+-0.2} & \num{1.63+-0.03} \\
    bw.\ loop time & \num{58.1+-1.1} & \num{2.38+-0.06} & \num{7.4+-0.3} & \num{7.6+-1.3} \\
    \midrule
    fw.\ time / step & \num{73+-3} & \num{62.1+-1.6} & \num{66.1+-1.6} & \num{60.9+-0.3} \\
    fw.\ model time / step & \num{71+-3} & \num{60.5+-1.5} & \num{62.6+-1.7} & \num{59.2+-0.2} \\
    bw.\ time / step & \num{3100+-300} & \num{555+-12} & \num{563+-8} & \num{540+-3} \\
    bw.\ model time / step & \num{3100+-300} & \num{553+-12} & \num{556+-9} & \num{532+-3} \\
    fw.\ steps & \num{13.4+-1.6} & \num{15+-1} & \num{16+-3} & \num{17+-3} \\
    bw.\ steps & \num{9+-1} & \num{12+-1} & \num{14+-4} & \num{13+-5} \\
    bits / dim & \num{1.38+-0.14} & \num{1.268+-0.015} & \num{1.28+-0.02} & \num{1.28+-0.03} \\
    \bottomrule
  \end{tabular}
\end{table}

The library versions we used are \pytorch{} \version{1.12.1}, \torchdiffeq{} \version{0.2.3} and \torchdyn{} \version{1.0.3} with an unreleased bug fix for the error estimate of the \texttt{dopri5} method. For \diffrax{} we used \version{0.2.1} and \jax{} \version{0.3.16}. All experiments used the 5th order Dormand-Prince method usually abbreviated \texttt{dopri5} \citep{dormand1980family} because it is consistently implemented across all libraries, even though the 5th order Tsitouras method \texttt{tsit5} \citep{tsitouras2011runge}, also available in \torchode{}, is often recommended over it today. For the same reason, we also opt for an integral controller in all experiments even though \diffrax{} and \torchode{} also implement a PID controller that could enable further performance gains (see \cref{sec:pid}).

We ran all benchmarks on an NVIDIA Geforce GTX 1080 Ti GPU with an Intel Xeon E5-2630 v4 CPU, because that is the most relevant configuration for deep learning applications and the situation that \torchode{} is optimized for. In particular, \torchode{}'s evaluation point tracking is implemented in a way that relies on the massive parallelism of a GPU.

In general, we measured the total time, the model time and the solver time per step for each setup. The total time measures everything that happens during a forward pass and is computed by measuring the total time for a prediction. Therefore it includes the time spent evaluating the model (learned dynamics), the time spent inside the solver itself as well as any surrounding code. Then we measure separately the time spent evaluating the (learned) model/dynamics and the time spent inside the \ODE{} solver (excluding the model time). The solver time divided by the number of solver steps is our main quantity of interest and we call it \emph{loop time}. Different solver implementations often take different numbers of steps for the same problem due to differing but equally valid implementation decisions. However, the time that each solver needs to make one step is independent of, for example, how exactly an internal error estimate is computed. Therefore, loop time is a fair and accurate metric to compare implementation efficiency across solvers.

All metrics and times are measured over three runs and are specified up to the first significant digit of the standard deviation; except, if that digit is \num{1}, we give an extra digit. \cref{table:benchmark} shows the mean loop times without standard deviations.

In the first benchmark, we solve a batch of \num{256} \vdp{} problems for one cycle with $\mu = 2$, absolute and relative tolerances of $10^{-5}$ and \num{200} evenly spaced evaluation points. Because evaluating the dynamics is so cheap in this case, we have not measured the model time separately for this setup and included it in the model time. Therefore, the loop time in \cref{table:benchmark-vdp} mostly measures how fast the solver can drive the GPU.\@ \torchode{} is then faster than \torchdiffeq{} and \torchdyn{} because it uses many combined \pytorch{} kernels and fewer tensor operations in total, which means that it can schedule the cheap dynamics evaluations faster. \jit{} compilation amplifies this effect by reducing the CPU overhead of the \python{} interpreter.

For the second benchmark, we have trained a \fen{} on the Black Sea dataset as in \citep{lienen2022learning} with batch size \num{8} and measure the times and metrics during the evaluation on the test set. First, we notice in \cref{table:benchmark-fen} that, again, \jit{} compilation reduces the loop time of \torchode{} significantly. Note that the learned dynamics are \jit{} compiled for all libraries, so this measures only the additional improvement from compiling the solver loop, too. Interestingly, \torchdyn{} is actually faster than non-compiled \torchode{} in this benchmark, in contrast to the previous benchmark. We suppose that this is because \torchdyn{}'s minimalistic implementation has less \python{} overhead than \torchode{} and because of the small number of evaluation points (\num{10}) and the smaller batch size compared to the \vdp{} benchmark, \torchode{}'s more efficient evaluation implementation carries less weight.

As a third benchmark, we repeat an experiment from \citep{grathwohl2019ffjord} and train a \cnf{} for density estimation on MNIST using the code accompanying their paper\footnote{\href{https://github.com/rtqichen/ffjord}{github.com/rtqichen/ffjord}}. The batch size is \num{500} in this case. See \cref{table:benchmark-cnf} for the results. In this case, there is no \jit{} compiled version of \torchode{} in the data, because custom extensions of \pytorch{}'s automatic differentiation are currently not supported by its \jit{} compiler. Since learning via the adjoint equation \citep{chen2018neural} has to be implemented as a custom gradient propagation method, it is incompatible with \jit{} compilation as of \pytorch{} \version{1.12.1}.

One should notice immediately, that, while \torchode{} has the fastest forward loop time, its backward loop time is the slowest by more than an order of magnitude. The reason is the interaction between the adjoint equation and \torchode{}'s independent parallel solving of \ODEs{}. The adjoint equation is an \ODE{}, just like the equation described by the learned model. Therefore, \torchode{} solves a separate adjoint equation for every instance in a batch to eliminate any interference between these separate and independent \ODEs{}. However, the adjoint equation is often significantly larger than the original \ODE{} because it has an additional variable for every parameter of the model. Let's say we are solving an \ODE{} with an initial state $y_{0} \in \mathbb{R}^{b \times f}$ with batch size $b$, $f$ features and a model $f_{\theta}, \theta \in \mathbb{R}^{p}$ with $p$ parameters. Then the adjoint equation in \torchdyn{} and \torchdiffeq{} has size $bf + p$, while \torchode{} will by default solve an equation with $b(f + p)$ variables.

The achieved MAE and bits / dim, respectively, in \cref{table:benchmark-fen,table:benchmark-cnf} show that this independent solving of \ODEs{} has no positive effect on the learning process or the performance metrics achieved. We suppose that the learned dynamics are usually simple enough to not be susceptible to the failure case shown in \cref{sec:batching}. On the contrary, jointly solving the adjoint equation seems to be beneficial for the learning process as evidenced by the higher bits / dim of \torchode{} in \cref{table:benchmark-cnf}. For this reason, \torchode{} includes a separate adjoint equation backward pass that solves the adjoint equation jointly on the whole batch, shown in the column \torchode{}-joint in \cref{table:benchmark-cnf}. This version has a significantly faster backward loop than \torchdiffeq{} and \torchdyn{} because at the larger \ODE{} size of $bf + p$ the saved operations from Horner's rule and combined kernels produce appreciable time savings. Furthermore, \torchode{} avoids any computations related to evaluating the solution at intermediate points if only the final solution is of interest as is the case for \cnfs{}.

\section{Example Code}\label{sec:code}

\begin{listing}
  \inputminted{python}{example.py}
  \caption{A code example solving a batch of \vdp{} problems with \torchode{}.}\label{lst:example}
\end{listing}

\cref{lst:example} shows a code example that solves a batch of \vdp{} problems with \torchode{}. The recorded solution statistics show how \torchode{} keeps track of separate step sizes, step acceptance and solver status for every instance. The number of function evaluations is the same for all problem instances even though they differ in their number of solver steps, because, in general, the dynamics have to be evaluated on a batch of the same size as the initial condition that got passed into the solver. So the model will continue to be evaluated on a problem instance until all problems in the batch have been solved, though these ``overhanging'' evaluations do not influence the result anymore.

\section{Impact of PID Control}\label{sec:pid}

\begin{figure}[h]
  \centering
  \includegraphics{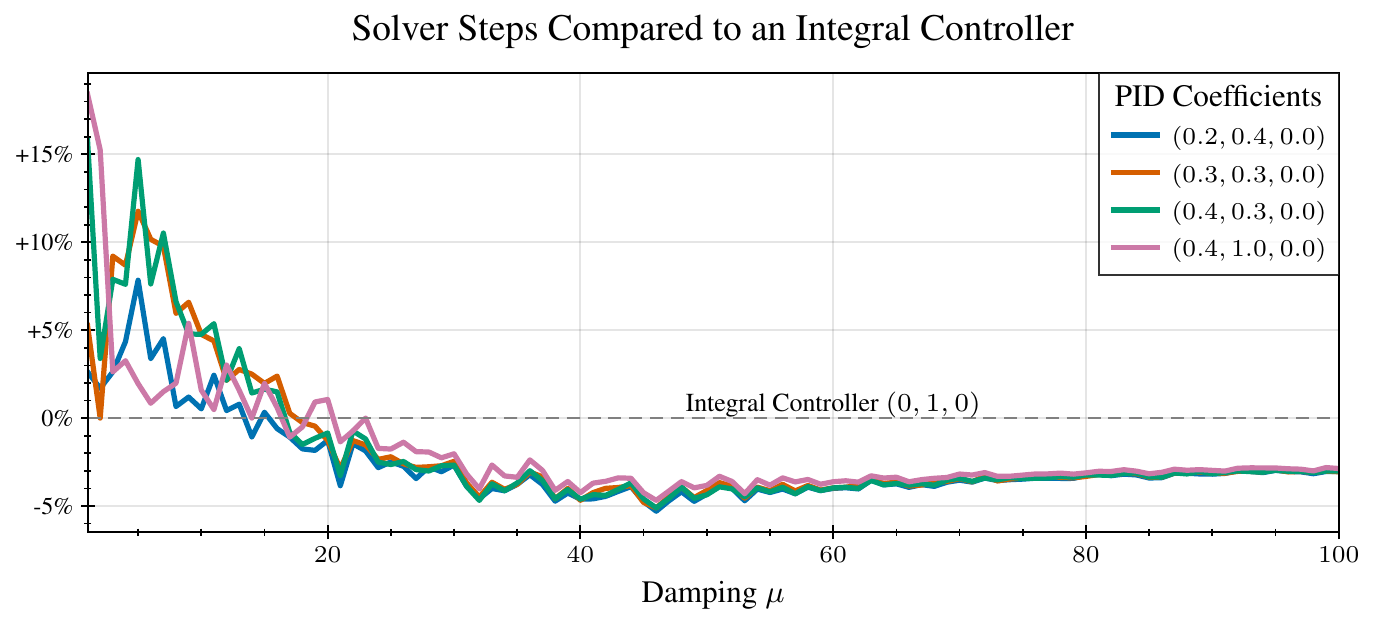}
  \caption{Solver steps required to solve one cycle of Van der Pol's oscillator (see \cref{eq:vdp}) with various PID coefficients compared to an integral controller.}\label{fig:pid}
\end{figure}
\vspace{1em}

A proportional-integral-derivative (PID) controller \citep{soderlind2002automatic,soderlind2003digital} can improve the performance of \ODE{} solvers by avoiding unnecessary solver steps when the stiffness of the problem changes abruptly. Its decision about growing or shrinking the next solver time step is based on the last three error estimates instead of only the current one as in an integral controller. This way it can react more accurately when the stability requirements on the time step change quickly as in \cref{fig:van-der-pol} and avoid rejected solver steps because of a too large step size as well as unnecessarily small solver steps.

To gain some insight into the effect of PID control on the number of solver steps, we solve Van der Pol's \cref{eq:vdp} for one cycle with various PID coefficients\footnote{We have taken the coefficients from \href{https://docs.kidger.site/diffrax/api/stepsize_controller/\#diffrax.PIDController}{diffrax's documentation}.} and compare the number of solver steps to the steps that the same solver would take with an integral controller. By varying the damping strength $\mu$ and therefore also the stiffness of the problem, we can control how strongly the step size varies across one cycle. See \cref{fig:van-der-pol} for the step sizes at $\mu = 25$. For $\mu = 0$, the limit cycle in phase space is a circle with very smooth step size behavior. With growing $\mu$, the limit cycle becomes more and more distorted and the variance in step size grows.

The results in \cref{fig:pid} show that there is a trade-off. For small variance in step size, i.e.\ $\mu < 15$, the PID controllers even take more steps than an integral controller. Only after $\mu > 25$ does PID control actually pay off with $3$ to $5\%$ in step savings over an integral controller.

We conclude that PID control is a valuable tool for \ODE{} problems that are difficult in the sense that the step size for an explicit method varies quickly and by at least two orders of magnitude. Given that the step size behavior of learned \ODE{} models is quite benign in our experience, we recommend the simple integral controller by default for deep learning applications and to try a PID controller when the number of solver steps exceeds $100$ or a significant variation in step size has been observed.

\end{document}